\documentclass[journal]{IEEEtran}

\usepackage{cite}
\usepackage[cmex10]{amsmath}
\usepackage{algorithmic}
\usepackage{array}
\usepackage{amsfonts,amssymb}
\usepackage{graphicx,float}
\usepackage{url,tabularx}
\usepackage{subfigure}
\usepackage{bm,amssymb}
\usepackage{accents}
\usepackage{amsthm}
\usepackage{comment}
\usepackage{color}
\usepackage{framed}
\usepackage{multicol}

\newcommand{\mb}[1]{\mathbf{#1}}
\newcommand{\mc}[1]{\mathcal{#1}}

\newcommand{\mbb}[1]{\mathbb{#1}}

\definecolor{lightgray}{gray}{0.95} 

\newcommand{\A}{\mb{A}}
\newcommand{\B}{\mb{B}}
\newcommand{\Bm}{\B_\mu}

\newcommand{\M}{\mb{M}}
\newcommand{\Id}{\mb{I}}

\newcommand{\wv}{\mb{w}}
\newcommand{\xv}{\mb{x}}
\newcommand{\hxv}{\widehat{\xv}}
\newcommand{\yv}{\mb{y}}
\newcommand{\zv}{\mb{z}}

\newcommand{\gv}{\mb{g}}

\newcommand{\uv}{\mb{u}}
\newcommand{\vv}{\mb{v}}
\newcommand{\delv}{\bm{\delta}}
\newcommand{\gamv}{\bm{\gamma}}
\newcommand{\epsv}{\bm{\epsilon}}
\newcommand{\byv}{\bar{\yv}}
\newcommand{\pv}{\mb{p}}
\newcommand{\dv}{\mb{d}}

\newcommand{\wk}{\wv_k}

\newcommand{\xk}{\xv_k}
\newcommand{\xkp}{\xv_{k+1}}
\newcommand{\xkm}{\xv_{k-1}}

\newcommand{\gk}{\gv_k}

\newcommand{\uk}{\uv_k}

\newcommand{\vk}{\vv_k}

\newcommand{\vkm}{\vv_{k-1}}
\newcommand{\delk}{\delv_k}
\newcommand{\delkp}{\delv_{k+1}}

\newcommand{\gamk}{\gamv_k}

\newcommand{\pk}{\pv_k}

\newcommand{\xkn}{\xv_{k_n}}
\newcommand{\xknp}{\xv_{k_n+1}}
\newcommand{\xknm}{\xv_{k_n-1}}
\newcommand{\wkn}{\wv_{k_n}}
\newcommand{\delkn}{\delv_{k_n}}
\newcommand{\ukn}{\uv_{k_n}}

\newcommand{\xd}{\xv_\bullet}
\newcommand{\xdd}{\xv_{\bullet\bullet}}
\newcommand{\ud}{\uv_\bullet}
\newcommand{\wvd}{\wv_\bullet}

\newcommand{\alpk}{\alpha_k}
\newcommand{\alpkn}{\alpha_{k_n}}
\newcommand{\etak}{\eta_k}

\newcommand{\F}{F}
\newcommand{\Qm}{Q_\mu}
\newcommand{\f}{f}
\newcommand{\gf}{\nabla f}
\newcommand{\pmu}[1]{\mc{P}_{#1}}
\newcommand{\tmu}[1]{\mc{H}_{#1}}

\newcommand{\g}{g}
\newcommand{\sgn}{\textup{sgn}}
\newcommand{\kn}{k_n}

\newcommand{\lam}{\lambda}
\newcommand{\Ind}{\mbb{I}}
\newcommand{\Phim}{\Phi_\mu}
\newcommand{\lamu}{\lam/\mu}
\newcommand{\gam}{\gamma}

\newcommand{\Ord}{\mc{O}}
\newcommand{\Zo}{\mc{Z}}
\newcommand{\Zc}{\mc{Z}^c}
\newcommand{\Fd}{\F_\bullet}
\newcommand{\sigk}{\sigma_k}
\newcommand{\eps}{\epsilon}
\newcommand{\phiZo}{\phi_{\Zo}}
\newcommand{\phiZc}{\phi_{\Zc}}

\begin{document}

\title{MIST: $\lowercase{l}_0$ Sparse Linear Regression with Momentum}

\author{Goran~Marjanovic,
          ~Magnus~O.~Ulfarsson, $^\dagger$
          ~Alfred~O.~Hero~III $^\ddagger$
\thanks{Goran Marjanovic is with the School of Electrical Engineering and Telecommunications, University of New South Wales, Sydney, NSW, 2052 Australia (email: goran.m.marjanovic@gmail.com)}
\thanks{Alfred O. Hero III is with the Department of Electrical and Computer Science, University of Michigan, Ann Arbor,
MI, 48109 USA (email: hero@eecs.umich.edu)}
\thanks{Magnus O. Ulfarsson is with the Department of Electrical Engineering, University of Iceland, Reykjavik, 107 Iceland (email: mou@hi.is)}
\thanks{$^\dagger$ This work was partly supported by the Research Fund of the University of Iceland and the Icelandic Research Fund (130635-051).}
\thanks{$^\ddagger$ This work was partially supported by ARO grant W911NF-11-1-0391.}
}

\maketitle


\begin{abstract}
Significant attention has been given to minimizing a penalized least squares criterion for estimating sparse solutions to large linear systems of equations. The penalty is responsible for inducing sparsity and the natural choice is the so-called $l_0$ norm. In this paper we develop a Momentumized Iterative Shrinkage Thresholding (MIST) algorithm for minimizing the resulting non-convex criterion and prove its convergence to a local minimizer. Simulations on large data sets show superior performance of the proposed method to other methods. 
\end{abstract}


\section{Introduction}

In the current age of big data acquisition there has been an ever growing interest in sparse representations, which consists of representing, say, a noisy signal as a linear combination of very few components. This implies that the entire information in the signal can be approximately captured by a small number of components, which has huge benefits in analysis, processing and storage of high dimensional signals. As a result, sparse linear regression has been widely studied with many applications in signal and image processing, statistical inference and machine learning. Specific applications include compressed sensing, denoising, inpainting, deblurring, source separation, sparse image reconstruction, and signal classification, etc. 

The linear regression model is given by:
\begin{align}
\yv=\A\xv + \epsv, \label{model}
\end{align}
where $\yv_{d\times 1}$ is a vector of noisy data observations, $\xv_{m\times 1}$ is the sparse representation (vector) of interest, $\A_{d\times m}$ is the regression matrix and $\epsv_{d\times 1}$ is the noise. The estimation aim is to choose the simplest model, i.e., the sparsest $\xv$, that adequately explains the data $\yv$. To estimate a sparse $\xv$, major attention has been given to minimizing a sparsity Penalized Least Squares (PLS) criterion \cite{BYD07,BD208,MFH11,BLR09,Tseng01,Nikolova13,MS13,MS14,BT209,WNF09}. The least squares term promotes goodness-of-fit of the estimator while the penalty shrinks its coefficients to zero. Here we consider the non-convex $l_0$ penalty since it is the natural sparsity promoting penalty and induces maximum sparsity. The resulting non-convex $l_0$ PLS criterion is given by: 
\begin{align}
\F(\xv)=\frac{1}{2}\|\yv-\A\xv\|_2^2+\lam\|\xv\|_0, \label{L0 PLS}
\end{align} 
where $\lam>0$ is the tuning/regularization parameter and $\|\xv\|_0$ is the $l_0$ penalty representing the number of non-zeros in $\xv$. 

\subsection{Previous Work}

Existing algorithms for directly minimizing (\ref{L0 PLS}) fall into the category of Iterative Shrinkage Thresholding (IST), and rely on the Majorization-Minimization (MM) type procedures, see \cite{BYD07,BD208}. These procedures exploit separability properties of the $l_0$ PLS criterion, and thus, rely on the minimizers of one dimensional versions of the PLS function: the so-called hard-thresholding operators. Since the convex $l_1$ PLS criterion has similar separability properties, some MM procedures developed for its minimization could with modifications be applied to minimize (\ref{L0 PLS}). Applicable MM procedures include first order methods and their accelerated versions \cite{BT209,BT309,FBDN07}. However, when these are applied to the $l_0$ penalized problem (\ref{L0 PLS}) there is no guarantee of convergence, and for \cite{BT209} there is additionally no guarantee of algorithm stability. 

Analysis of convergence of MM algorithms for minimizing the $l_0$ PLS criterion (\ref{L0 PLS}) is rendered difficult due to lack of convexity. As far as we are aware, algorithm convergence for this problem has only been shown for the Iterative Hard Thresholding (IHT) method \cite{BYD07,BD208}. Specifically, a bounded sequence generated by IHT was shown to converge to the set of local minimizers of (\ref{L0 PLS}) when the singular values of $\A$ are strictly less than one. Convergence analysis of algorithms designed for minimizing the $l_q$ PLS criterion, $q\in(0,1]$, is not applicable to the case of the $l_0$ penalized objective (\ref{L0 PLS}) because it relies on convex arguments when $q=1$, and continuity and/or differentiability of the criterion when $q\in(0,1)$. 

\subsection{Paper Contribution}

In this paper we develop an MM algorithm with momentum acceleration, called Momentumized IST (MIST), for minimizing the $l_0$ PLS criterion (\ref{L0 PLS}) and prove its convergence to a single local minimizer without imposing any assumptions on $\A$. Simulations on large data sets are carried out, which show that the proposed algorithm outperforms existing methods for minimizing (\ref{L0 PLS}), including modified MM methods originally designed for the $l_1$ PLS criterion. 

The paper is organised as follows. Section \ref{preliminaries section} reviews some of background on MM that will be used to develop the proposed convergent algorithm. The proposed algorithm is given in Section \ref{the algorithm section}, and Section \ref{convergence section} contains the convergence analysis. Lastly, Section \ref{simulations section} and \ref{conclusion section} presents the simulations and concluding remarks respectively. 

\subsection{Notation}

The $i$-th component of a vector $\vv$ is denoted by $\vv[i]$. Given a vector $f(\vv)$, where $f(\cdot)$ is a function, its $i$-th component is denoted by $f(\vv)[i]$. $\|\M\|$ is the spectral norm of matrix $\M$. $\Ind(\cdot)$ is the indicator function equaling $1$ if its argument is true, and zero otherwise. Given a vector $\vv$, $\|\vv\|_0=\sum_i\Ind(\vv[i]\neq 0)$. $\sgn(\cdot)$ is the sign function. $\{\xk\}_{k\geq 0}$ denotes an infinite sequence, and $\{\xkn\}_{n\geq 0}$ an infinite subsequence, where $\kn\leq k_{n+1}$ for all $n\geq 0$.

\section{Preliminaries} \label{preliminaries section}

Denoting the least squares term in (\ref{L0 PLS}) by:
\begin{align}
f(\xv)=\frac{1}{2}\|\yv-\A\xv\|_2^2, \nonumber
\end{align}
the Lipschitz continuity of $\gf(\cdot)$ implies:
\begin{equation}
\f(\zv)\leq\f(\xv)+\gf(\xv)^T(\zv-\xv)+\frac{\mu}{2}\|\zv-\xv\|_2^2 \nonumber
\end{equation}
for all $\xv,\zv,\mu\geq\|\A\|^2$. For the proof see \cite[Lemma 2.1]{BT209}. As a result, the following approximation of the objective function $\F(\cdot)$ in (\ref{L0 PLS}),
\small
\begin{align}
\Qm(\zv,\xv)=\f(\xv)+\gf(\xv)^T(\zv-\xv)+\frac{\mu}{2}\|\zv-\xv\|_2^2 + \lam\|\zv\|_0 \label{Q function}
\end{align} 
\normalsize 
is a majorizing function, i.e.,
\begin{align}
\F(\zv)\leq\Qm(\zv,\xv) \textup{ for any }\xv,\zv,\mu\geq\|\A\|^2. \label{MM inequality}
\end{align}
Let $\pmu{\mu}(\xv)$ be any point in the set $\arg\min_{\zv}\Qm(\zv,\xv)$, we have:
\begin{equation}
\F(\pmu{\mu}(\xv))\overset{(\ref{MM inequality})}{\leq}\Qm(\pmu{\mu}(\xv),\xv)\leq\Qm(\xv,\xv)=\F(\xv), \label{Fxkp<=Fxk}
\end{equation}
where the stacking of (\ref{MM inequality}) above the first inequality indicates that this inequality follows from Eq. (\ref{MM inequality}). The proposed algorithm will be constructed based on the above MM framework with a momentum acceleration, described below. This momentum acceleration will be designed based on the following.
\newtheorem{L0 FISTA theorem}{Theorem}
\begin{L0 FISTA theorem} \label{L0 FISTA thm}
\textup{
Let $\Bm=\mu\Id-\A^T\A$, where $\mu>\|\A\|^2$, and:
\begin{align}
\alpha=2\eta\left(\frac{\delv^T\Bm(\pmu{\mu}(\xv)-\xv)}{\delv^T\Bm\delv}\right),\ \eta\in[0,1], \label{alpk for L0 FISTA}
\end{align}
where $\delv\neq 0$. Then, $\F\left(\pmu{\mu}(\xv+\alpha\delv)\right)\leq\F(\xv)$.
}
\end{L0 FISTA theorem}

For the proof see the Appendix. 

\subsection{Evaluating the Operator $\pmu{\mu}(\cdot)$}

Since (\ref{Q function}) is non-convex there may exist multiple minimizers of $\Qm(\zv,\cdot)$ so that $\pmu{\mu}(\cdot)$ may not be unique. We select a single element of the set of minimizers as described below. By simple algebraic manipulations of the quadratic quantity in (\ref{Q function}), letting:
\begin{align}
\g(\xv)=\xv-\frac{1}{\mu}\gf(\xv), \label{g function}
\end{align}
it is easy to show that:
\begin{align}
\Qm(\zv,\xv)=\frac{\mu}{2}\|\zv-\g(\xv)\|_2^2+\lam\|\zv\|_0+\f(\xv)-\frac{1}{2\mu}\|\gf(\xv)\|_2^2, \nonumber
\end{align}
and so, $\pmu{\mu}(\cdot)$ is given by:
\begin{align}
\pmu{\mu}(\xv)=\arg\min_{\zv}\ \frac{1}{2}\left\|\zv-\g(\xv)\right\|_2^2+(\lamu)\|\zv\|_0. \label{pmu operator}
\end{align}
For the proposed algorithm we fix $\pmu{\mu}(\cdot)=\tmu{\lam/\mu}(\g(\cdot))$, the point to point map defined in the following Theorem. 
\newtheorem{hard-thresholding theorem}[L0 FISTA theorem]{Theorem}
\begin{hard-thresholding theorem} \label{hard-thresholding thm}
\textup{
Let the hard-thresholding (point-to-point) map $\tmu{h}(\cdot)$, $h>0$, be such that for each $i=1,\dots,m$:
\small
\begin{align}
\tmu{h}(\g(\vv))[i]=
\begin{cases}
0 & \textup{if }|\g(\vv)[i]|<\sqrt{2h} \\[1mm]
\g(\vv)[i]\ \Ind(\vv[i]\neq 0) & \textup{if }|\g(\vv)[i]|=\sqrt{2h} \\[1mm]
\g(\vv)[i] & \textup{if }|\g(\vv)[i]|>\sqrt{2h}.
\end{cases} \label{H map}
\end{align}
\normalsize 
Then, $\tmu{\lamu}(\g(\vv))\in\arg\min_{\zv}\Qm(\zv,\vv)$, where $\g(\cdot)$ is from (\ref{g function}).
}
\end{hard-thresholding theorem}

The proof is in the Appendix. 

\vspace{2mm}

Evidently Theorem \ref{L0 FISTA thm} holds with $\pmu{\mu}(\cdot)$ replaced by $\tmu{\lamu}(\g(\cdot))$. The motivation for selecting this particular minimizer is Lemma \ref{closed map thm} in Section \ref{convergence section}.  

\section{The Algorithm} \label{the algorithm section}

The proposed algorithm is constructed by repeated application of Theorem \ref{L0 FISTA thm} where $\delv$ is chosen to be the difference between the current and the previous iterate, i.e.,
\small
\begin{align}
\xkp=\tmu{\lamu}\left(\wk-\frac{1}{\mu}\gf(\wk)\right),\ \wk=\xk+\alpk\delk \label{iteration formula}
\end{align}
\normalsize
with $\alpk$ given by (\ref{alpk for L0 FISTA}), where $\delk=\xk-\xkm$. The iteration (\ref{iteration formula}) is an instance of a momentum accelerated IST algorithm, similar to Fast IST Algorithm (FISTA) introduced in \cite{BT209} for minimizing the convex $l_1$ PLS criterion. In (\ref{iteration formula}), $\delk$ is called the momentum term and $\alpk$ is a momentum step size parameter. A more explicit implementation of (\ref{iteration formula}) is given below. Our proposed algorithm will be called Momentumized Iterative Shrinkage Thresholding (MIST).

\noindent\rule{\columnwidth}{0.07cm}\\*[-0.05cm]
\textbf{Momentumized IST (MIST) Algorithm} \\*[-0.25cm]
\rule{\columnwidth}{0.035cm}\\[1mm]
Compute $\byv=(\yv^T\A)^T$ off-line. Choose $\xv_0$ and let $\xv_{-1}=\xv_0$. Also, calculate $\|\A\|^2$ off-line, let $\mu>\|\A\|^2$ and $k=0$. Then:
\begin{enumerate}
\item[(1)] If $k=0$, let $\alpk=0$. Otherwise, compute: \\[1mm]
(a) $\uk=\A\xk$\\[2mm]
(b) $\vk=(\uk^T\A)^T$\\[2mm]
(c) $\gk=\xk-\frac{1}{\mu}(\vk-\byv)$\\[2mm]
(d) $\pk=\tmu{\lamu}(\gk)-\xk$\\[2mm]
(e) $\delk=\xk-\xkm$ and:
\begin{equation}
\gamk=\mu\delk-\vk+\vkm \label{Bdelv computed}
\end{equation}
(f) Choose $\etak\in(0,1)$ and compute:
\begin{align}
\alpk= 2\etak\left(\frac{\gamk^T\pk}{\gamk^T\delk}\right) \label{alpk computed}
\end{align}
\item[(2)] Using (c), (e) and (f) compute:
\begin{equation}
\xkp=\tmu{\lamu}\left(\gk+\frac{\alpk}{\mu}\gamk\right) \label{xvkp computed}
\end{equation}
\item[(3)] Let $k=k+1$ and go to (1).
\end{enumerate}
\vspace{-2mm}
\rule{\columnwidth}{0.07cm}

\newtheorem{complexity remark}{Remark}
\begin{complexity remark} \label{complexity rmk}
\textup{
Thresholding using (\ref{H map}) is simple, and can always be done off-line. Secondly, note that MIST requires computing only $\Ord(2md)$ products, which is the same order required when the momentum term $\delk$ is not incorporated, i.e., $\etak=0$ for all $k$. In this case, MIST is a generalization of IHT from \cite{BYD07,BD208}. Other momentum methods such as FISTA \cite{BT209} and its monotone version M-FISTA \cite{BT309} also require computing $\Ord(2md)$ and $\Ord(3md)$ products, respectively.
}
\end{complexity remark}

\section{Convergence Analysis} \label{convergence section}

Here we prove that the proposed MIST algorithm iterates converge to a local minimizer of $\F(\cdot)$.

\newtheorem{convergence theorem}[L0 FISTA theorem]{Theorem}
\begin{convergence theorem} \label{convergence thm}
\textup{Suppose $\{\xk\}_{k\geq 0}$ is a bounded sequence generated by the MIST algorithm. Then $\xk\to\xd$ as $k\to\infty$, where $\xd$ is a local minimizer of (\ref{L0 PLS}). 
}
\end{convergence theorem} 

The proof is in the Appendix and requires several lemmas that are also proved in the Appendix. 

In Lemma \ref{difference in iterates to zero lem} and \ref{closed map thm} it is assumed that MIST reaches a fixed point only in the limit, i.e., $\xkp\neq\xk$ for all $k$. This implies that $\delk\neq 0$ for all $k$.

\newtheorem{difference in iterates to zero lemma}{Lemma}
\begin{difference in iterates to zero lemma} \label{difference in iterates to zero lem}
\textup{
$\xkp-\xk\to 0$ as $k\to\infty$.
}
\end{difference in iterates to zero lemma} 

The following lemma motivates Theorem \ref{hard-thresholding thm} and is crucial for the subsequent convergence analysis.

\newtheorem{closed map theorem}[difference in iterates to zero lemma]{Lemma}
\begin{closed map theorem} \label{closed map thm}
\textup{
Assume the result in Lemma \ref{difference in iterates to zero lem}. If, for any subsequence $\{\xkn\}_{n\geq 0}$, $\xkn\to\xd$ as $n\to\infty$, then:
\small
\begin{align}
\tmu{\lamu}\left(\wkn-\frac{1}{\mu}\gf(\wkn)\right) \to \tmu{\lamu}\left(\xd-\frac{1}{\mu}\gf(\xd)\right), \label{H is closed}
\end{align}
\normalsize
where $\wkn=\xkn+\alpkn\delkn$.  
}
\end{closed map theorem} 

The following lemma characterizes the fixed points of the MIST algorithm.

\newtheorem{fixed points lemma}[difference in iterates to zero lemma]{Lemma}
\begin{fixed points lemma} \label{fixed points lem}
\textup{
Suppose $\xd$ is a fixed point of MIST. Letting $\Zo=\{i:\xd[i]=0\}$ and $\Zc=\{i:\xd[i]\neq 0\}$,
\vspace{1mm}
\begin{itemize}
\item[(C$_1$)] If $i\in\Zo$, then $|\gf(\xd)[i]|\leq\sqrt{2\lam\mu}$.
\vspace{1mm}
\item[(C$_2$)] If $i\in\Zc$, then $\gf(\xd)[i]=0$.
\vspace{1mm}
\item[(C$_3$)] If $i\in\Zc$, then $|\xd[i]|\geq\sqrt{2\lamu}$.
\end{itemize}
}
\end{fixed points lemma} 

\newtheorem{fixed points are local min lemma}[difference in iterates to zero lemma]{Lemma}
\begin{fixed points are local min lemma} \label{fixed points are local min lem}
\textup{
Suppose $\xd$ is a fixed point of MIST. Then there exists $\eps>0$ such that $\F(\xd)<\F(\xd+\dv)$ for any $\dv$ satisfying $\|\dv\|_2\in(0,\eps)$. In other words, $\xd$ is a strict local minimizer of (\ref{L0 PLS}).
}
\end{fixed points are local min lemma}

\newtheorem{limit points are fixed points lemma}[difference in iterates to zero lemma]{Lemma}
\begin{limit points are fixed points lemma} \label{limit points are fixed points lem}
\textup{
The limit points of $\{\xk\}_{k\geq 0}$ are fixed points of MIST.
}
\end{limit points are fixed points lemma}

All of the above lemmas are proved in the Appendix.

\section{Simulations} \label{simulations section}

Here we demonstrate the performance advantages of the proposed MIST algorithm in terms of convergence speed. The methods used for comparison are the well known MM algorithms: ISTA and FISTA from \cite{BT209}, as well as M-FISTA from \cite{BT309}, where the soft-thresholding map is replaced by the hard-thresholding map. In this case, ISTA becomes identical to the IHT algorithm from \cite{BYD07,BD208}, while FISTA and M-FISTA become its accelerated versions, which exploit the ideas in \cite{Nesterov83}. 

\begin{figure}[!ht]
\centering
\begin{minipage}{0.6\linewidth}
  \centering
  \includegraphics[width=1.0\linewidth]{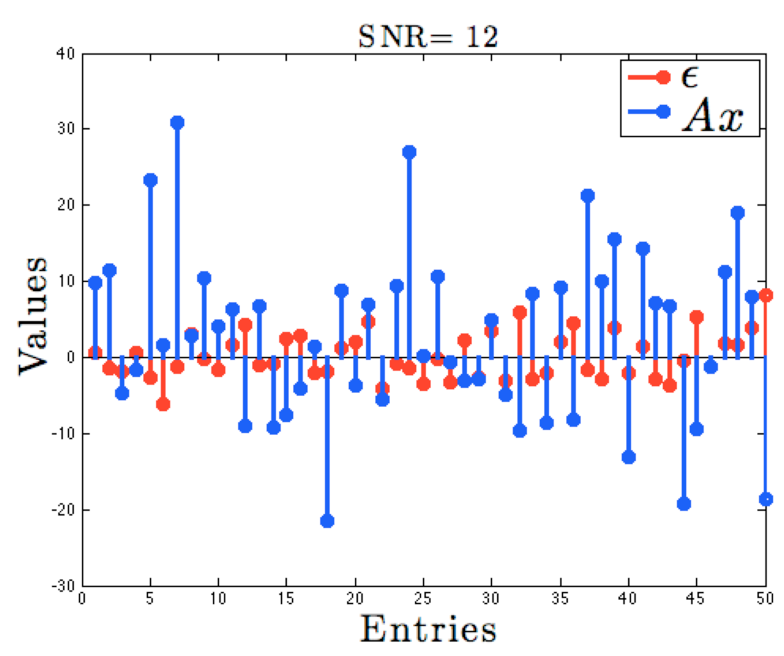}
\end{minipage}%
\caption{Examples of the first $50$ entry values in the (noiseless) observation $\A\xv$ and the observation noise $\epsv$ for SNR$=12$.}
\label{SNR12p0 fig}

\vspace{2mm}

\centering
\begin{minipage}{0.6\linewidth}
  \centering
  \includegraphics[width=1.0\linewidth]{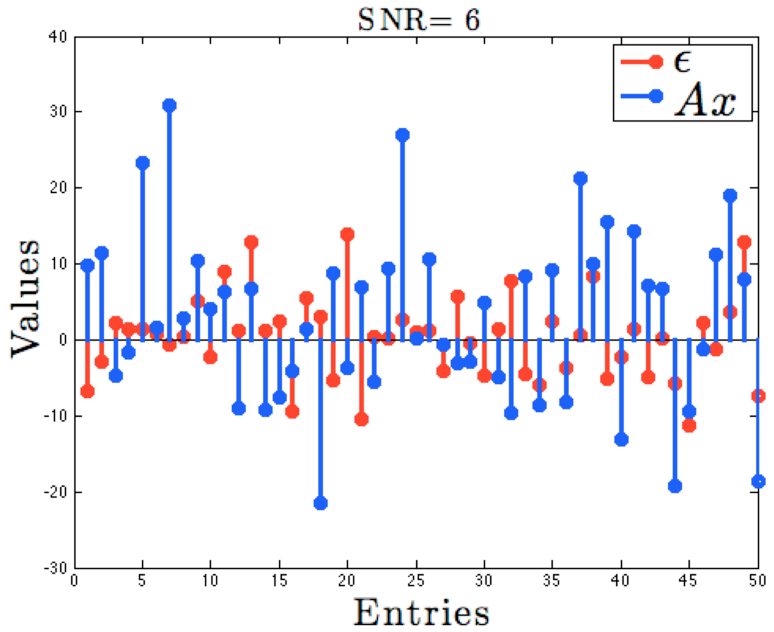}
\end{minipage}%
\caption{Examples of the first $50$ entry values in the (noiseless) observation $\A\xv$ and the observation noise $\epsv$ for SNR$=6$.}
\label{SNR6p0 fig}

\vspace{2mm}

\centering
\begin{minipage}{0.6\linewidth}
  \centering
  \includegraphics[width=1.0\linewidth]{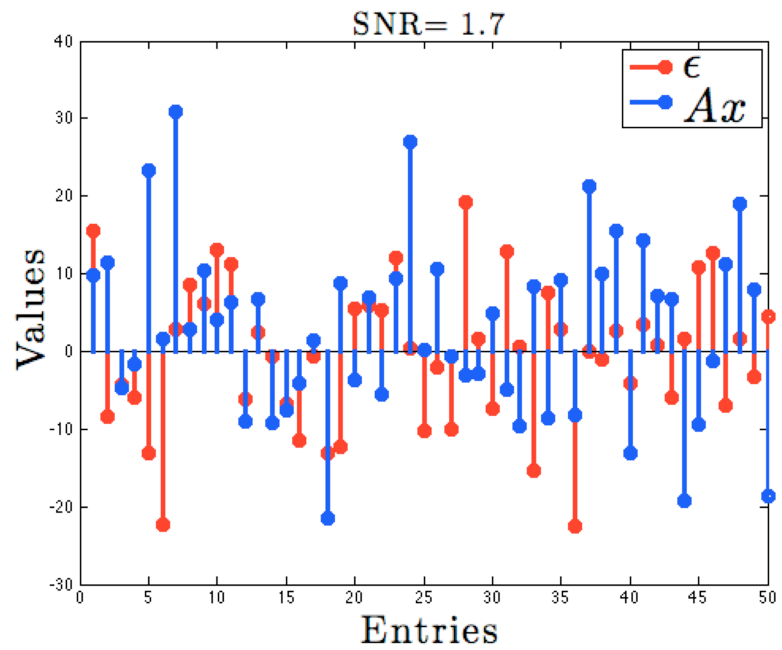}
\end{minipage}%
\caption{Examples of the first $50$ entry values in the (noiseless) observation $\A\xv$ and the observation noise $\epsv$ for SNR$=1.7$.}
\label{SNR1p7 fig}
\end{figure}

A popular compressed sensing scenario is considered with the aim of reconstructing a length $m$ sparse signal $\xv$ from $d$ observations, where $d<m$. The matrix $\A_{d\times m}$ is obtained by filling it with independent samples from the standard Gaussian distribution. A relatively high dimensional example is considered, where $d=2^{13}=8192$ and $m=2^{14}=16384$, and $\xv$ contains $150$ randomly placed $\pm 1$ spikes ($0.9\%$ non-zeros). The observation $\yv$ is generated according to (\ref{model}) with the standard deviation of the noise $\epsv$ given by $\sigma=3,6,10$. The Signal to Noise Ratio (SNR) is defined by:
\begin{align}
\textup{SNR}=10\log_{10}\left(\frac{\|\A\xv\|_2^2}{\sigma^2d}\right). \nonumber
\end{align} 
Figures \ref{SNR12p0 fig},  \ref{SNR6p0 fig} and \ref{SNR1p7 fig} show a plot of $\A\xv$ and observation noise $\epsv$ for the three SNR values corresponding to the three considered values of $\sigma$.

\subsection{Selection of the Tuning Parameter $\lam$}

Oracle results are reported, where the chosen tuning parameter $\lam$ in (\ref{L0 PLS}) is the minimizer of the Mean Squared Error (MSE), defined by:
\begin{align}
\textup{MSE}(\lam)=\frac{\|\xv-\hxv\|_2^2}{\|\xv\|_2^2}, \nonumber
\end{align}
where $\hxv=\hxv(\lam)$ is the estimator of $\xv$ produced by a particular algorithm. 

As $\xv$ is generally unknown to the experimenter we also report results of using a model selection method to select $\lam$. Some of the classical model selection methods include the Bayesian Information Criterion (BIC) \cite{Schwarz78}, the Akaike Information criterion \cite{Akaike73}, and (generalized) cross validation \cite{Stone74,CW79}. However, these methods tend to select a model with many spurious components when $m$ is large and $d$ is comparatively smaller, see \cite{CC08,BS02,S04,BDG04}. As a result, we use the Extended BIC (EBIC) model selection method proposed in \cite{CC08}, which incurs a small loss in the positive selection rate while tightly controlling the false discovery rate, a desirable property in many applications. The EBIC is defined by:
\small
\begin{align}
\textup{EBIC}(\lam)=\log\left(\frac{\|\yv-\A\hxv\|_2^2}{d}\right)+\left(\frac{\log d}{d}+2\gam\frac{\log m}{d}\right)\|\hxv\|_0, \nonumber
\end{align} 
\normalsize
and the chosen $\lam$ in (\ref{L0 PLS}) is the minimizer of this criterion. Note that the classical BIC criterion is obtained by setting $\gam=0$. As suggested in \cite{CC08}, we let $\gam=1-1/(2\kappa)$, where $\kappa$ is the solution of $m=d^\kappa$, i.e., 
\begin{align}
\kappa=\frac{\log m}{\log d}=1.08 \nonumber
\end{align}

\subsection{Results}

All algorithms are initialized with $\xv_0=\mb{0}$, and are terminated when the following criterion is satisfied:
\begin{align}
\frac{|\F(\xk)-\F(\xkm)|}{\F(\xk)}<10^{-10}. \label{termination criterion}
\end{align}
In the MIST algorithm we let $\mu=\|\A\|^2+10^{-15}$ and $\etak=1-10^{-15}$. All experiments were run in MATLAB 8.1 on an Intel Core i$7$ processor with 3.0GHz CPU and 8GB of RAM. 

\begin{figure}[!ht]
\centering
\begin{minipage}{1.0\linewidth}
  \centering
  \includegraphics[width=1.0\linewidth]{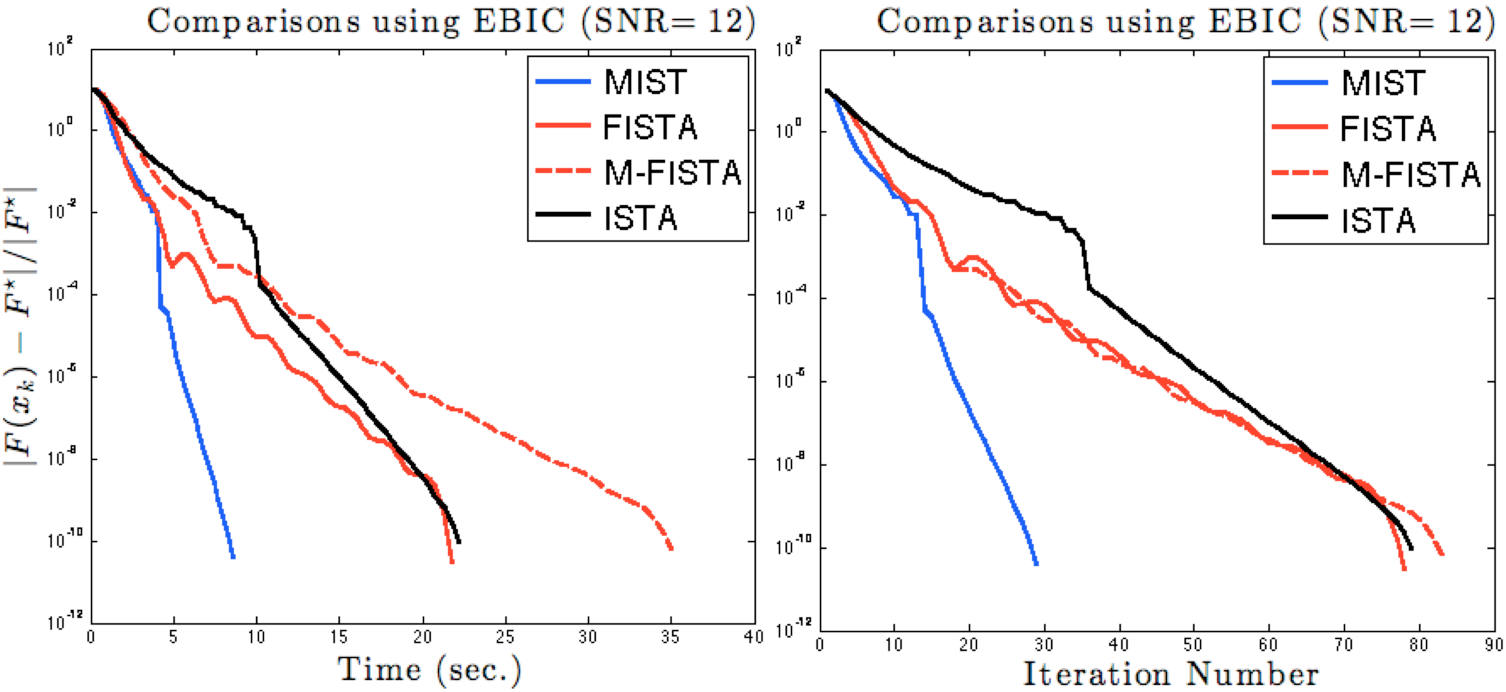}
\end{minipage}%
\caption{Algorithm comparisons based on relative error $|\F(\xk)-\F^\star|/|\F^\star|$ where $\F^\star$ is the final value of $\F(\cdot)$ obtained by each algorithm at its termination, i.e., $\F^\star=\F(\xk)$, where $\F(\xk)$ satisfies the termination criterion in (\ref{termination criterion}). Here SNR=12, and the regularization parameter $\lambda$ has been selected using the EBIC criterion. As it can be seen, in the low noise environment $(\sigma=3)$ the MIST algorithm outperforms the rest, both in terms of time and iteration number.}
\label{CompEBICSig03 fig}
\centering
\vspace{3mm}
\begin{minipage}{1.0\linewidth}
  \centering
  \includegraphics[width=1.0\linewidth]{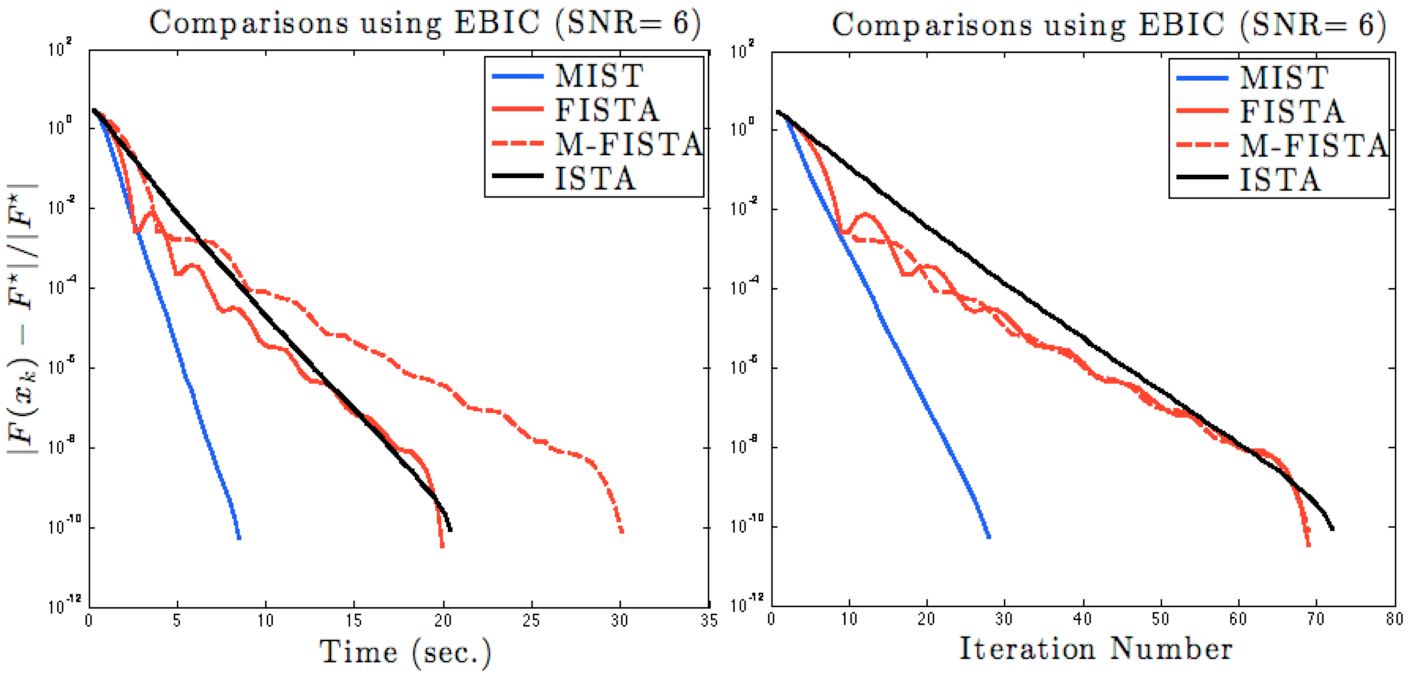}
\end{minipage}%
\caption{Similar comparisons as in Fig. \ref{CompEBICSig03 fig} except that SNR=6. As it can be seen, in the intermediate noise environment $(\sigma=6)$ the MIST algorithm outperforms the others, both in terms of time and iteration number.}
\label{CompEBICSig06 fig}
\centering
\vspace{3mm}
\begin{minipage}{1.0\linewidth}
  \centering
  \includegraphics[width=1.0\linewidth]{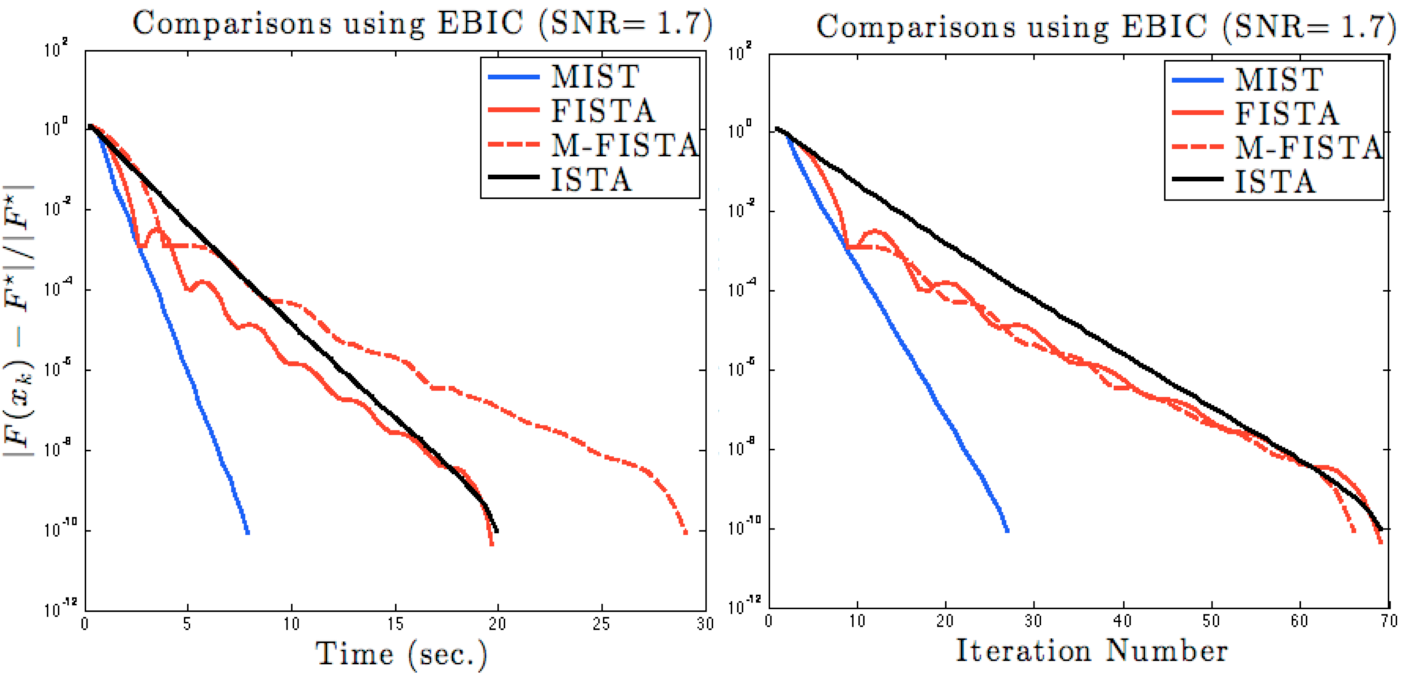}
\end{minipage}%
\caption{Similar comparisons as in Fig. \ref{CompEBICSig03 fig} except that SNR=1.7. As it can be seen, in the high noise environment $(\sigma=10)$ the MIST algorithm outperforms the rest, both in terms of time and iteration number.}
\label{CompEBICSig10 fig}
\end{figure}

Figures \ref{CompEBICSig03 fig}, \ref{CompEBICSig06 fig} and \ref{CompEBICSig10 fig} show percentage reduction of $\F(\cdot)$ as a function of time and iteration for each algorithm. To make the comparisons fair, i.e., to make sure all the algorithms minimize the same objective function, a common $\lam$ is used and chosen to be the smallest $\lam$ from the averaged $\arg\min_\lam\textup{EBIC}(\lam)$ obtained by each algorithm (over $10$ instances). 

\begin{figure}[!ht]
\centering
\begin{minipage}{1.0\linewidth}
  \centering
  \includegraphics[width=1.0\linewidth]{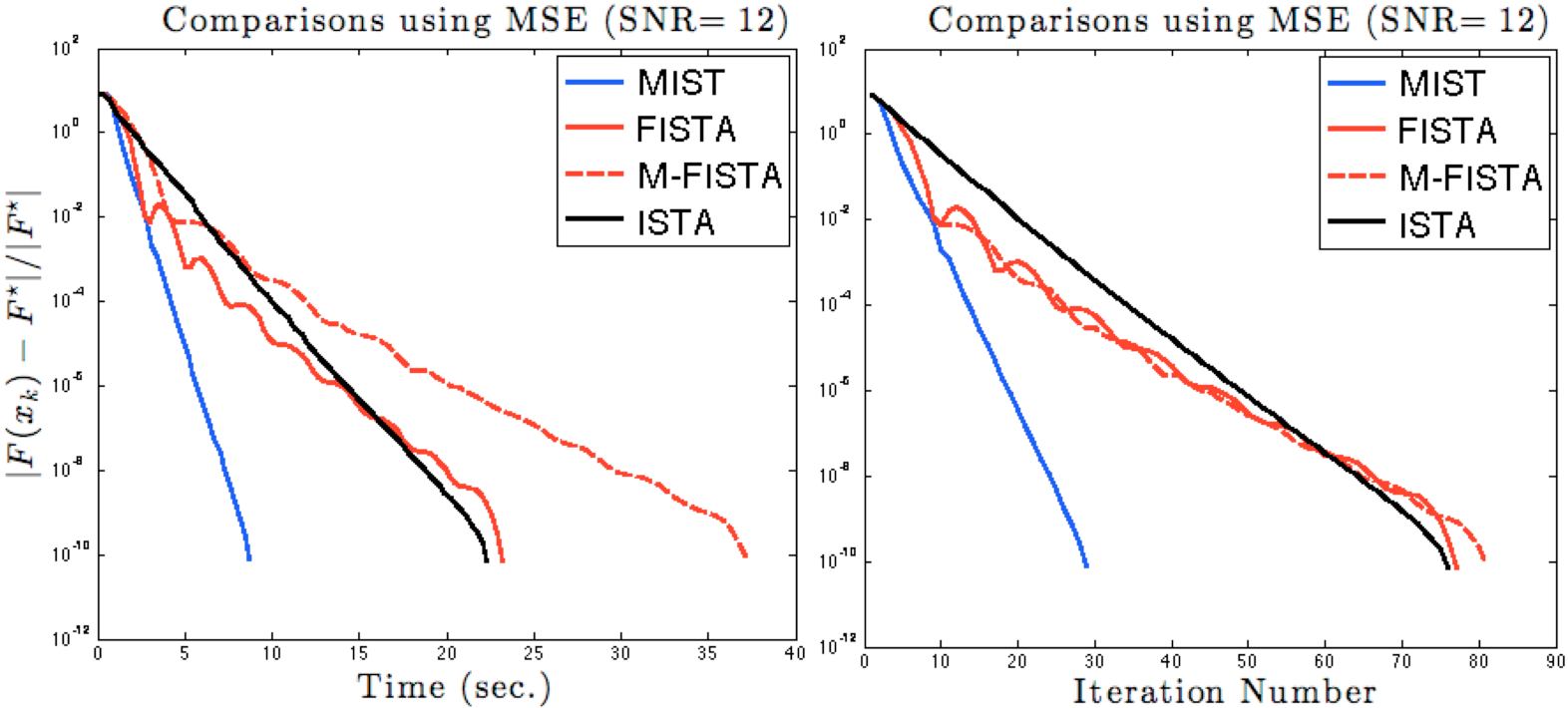}
\end{minipage}%
\caption{Algorithm comparisons based on relative error $|\F(\xk)-\F^\star|/|\F^\star|$ where $\F^\star$ is the final value of $\F(\cdot)$ obtained by each algorithm at its termination, i.e., $\F^\star=\F(\xk)$, where $\F(\xk)$ satisfies the termination criterion in (\ref{termination criterion}). Here an oracle selects the regularization parameter $\lambda$ using the minimum MSE criterion. As it can be seen, in the low noise environment $(\sigma=3)$ the MIST algorithm outperforms the rest, both in terms of time and iteration number.}
\label{CompMSESig03 fig}
\centering
\vspace{3mm}
\begin{minipage}{1.0\linewidth}
  \centering
  \includegraphics[width=1.0\linewidth]{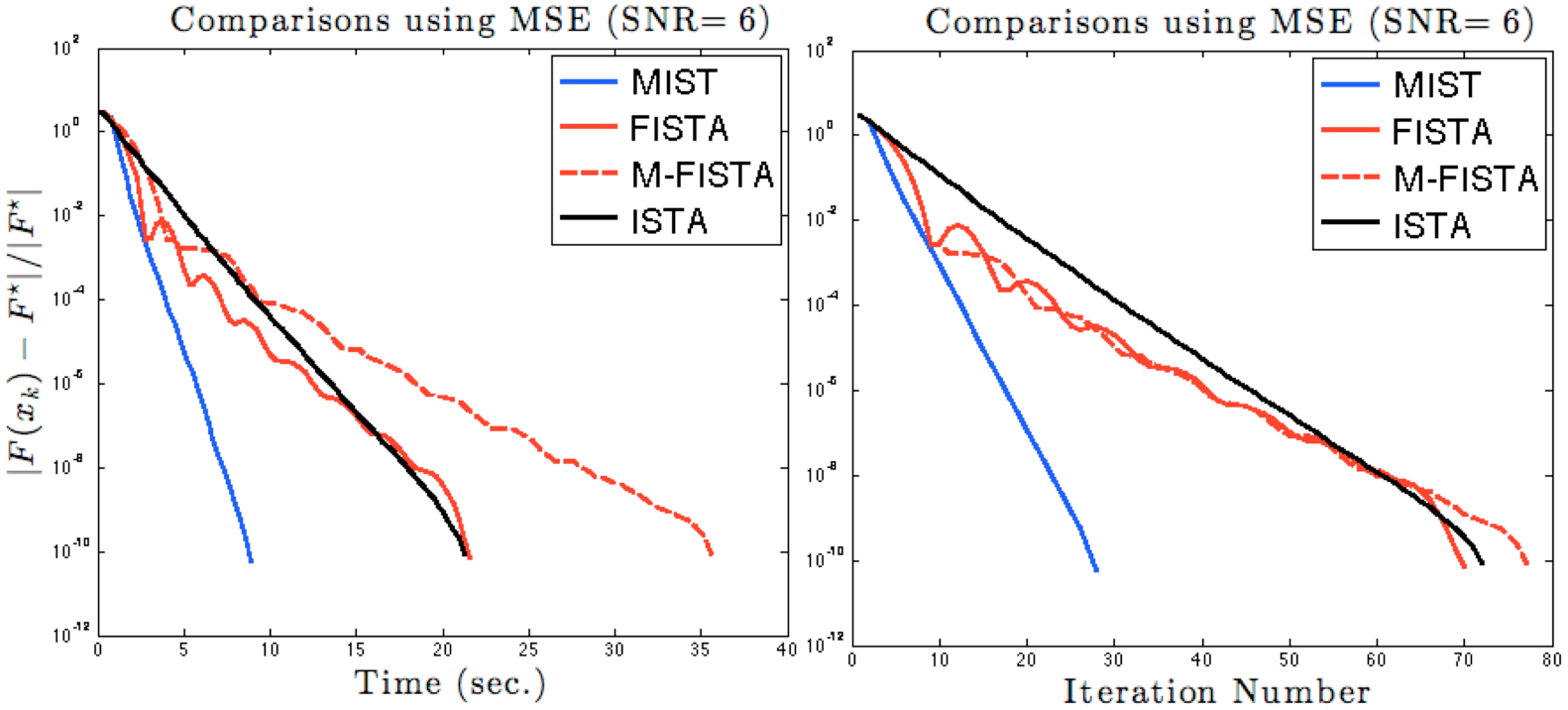}
\end{minipage}%
\caption{Similar comparisons as in Fig. \ref{CompMSESig03 fig} except that SNR=6. As it can be seen, in the intermediate noise environment $(\sigma=6)$ the MIST algorithm outperforms the rest, both in terms of time and iteration number.}
\label{CompMSESig06 fig}
\centering
\vspace{3mm}
\begin{minipage}{1.0\linewidth}
  \centering
  \includegraphics[width=1.0\linewidth]{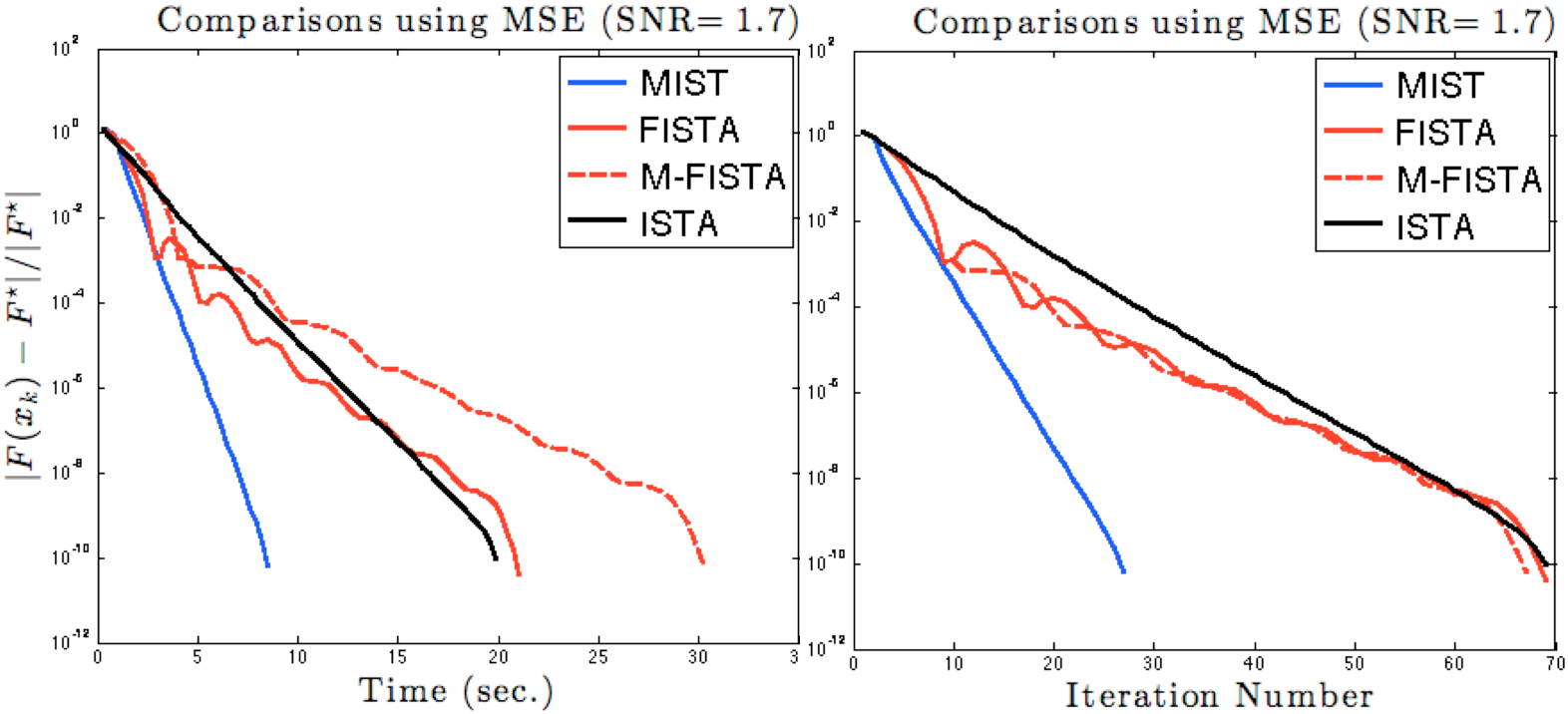}
\end{minipage}%
\caption{Similar comparisons as in Fig. \ref{CompMSESig03 fig} except that SNR=1.7. As it can be seen, in the high noise environment $(\sigma=10)$ the MIST algorithm outperforms the rest, both in terms of time and iteration number.}
\label{CompMSESig10 fig}
\end{figure}

Figures \ref{CompMSESig03 fig}, \ref{CompMSESig06 fig} and \ref{CompMSESig10 fig} also show percentage reduction of $\F(\cdot)$ as a function of time and iteration for each algorithm. This time the MSE is used for the tuning of $\lam$, and again, to make sure all the algorithms minimize the same objective function, a common $\lam$ is used and chosen to be the smallest $\lam$ from the averaged $\arg\min_\lam\textup{MSE}(\lam)$ obtained by each algorithm (over $10$ instances). 

Based on a large number of experiments we noticed that MIST, FISTA and ISTA outperformed M-FISTA in terms of run time. This could be due to the fact that M-FISTA requires computing a larger number of products, see Remark \ref{complexity rmk}, and the fact that it is a monotone version of a severely non-monotone FISTA. The high non-monotonicity could possibly be due to non-convexity of the objective function $\F(\cdot)$.   

\begin{figure}[!ht]
\centering
\begin{minipage}{0.7\linewidth}
  \centering
  \includegraphics[width=1.0\linewidth]{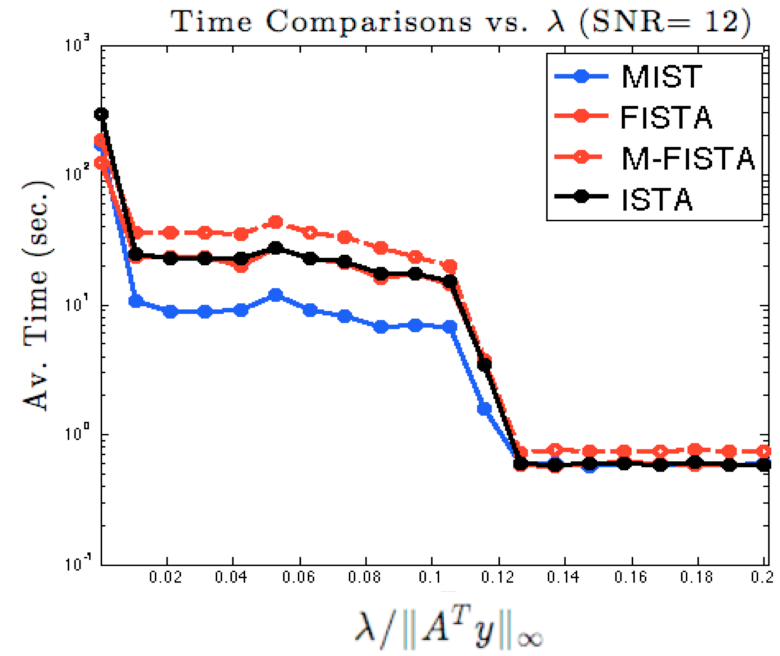}
\end{minipage}%
\caption{Average time vs. $\lam$ over $5$ instances when SNR=12. For the comparison, $20$ equally spaced values of $\lam$ are considered, where $10^{-4}\|\A^T\yv\|_\infty\leq\lam\leq0.2\|\A^T\yv\|_\infty$. As it can be seen, except in the scenario when $\lam=10^{-4}\|\A^T\yv\|_\infty$ the MIST algorithm outperforms the others.}
\label{LamVsTimeSNR12p0 fig}

\vspace{2mm}

\centering
\begin{minipage}{0.7\linewidth}
  \centering
  \includegraphics[width=1.0\linewidth]{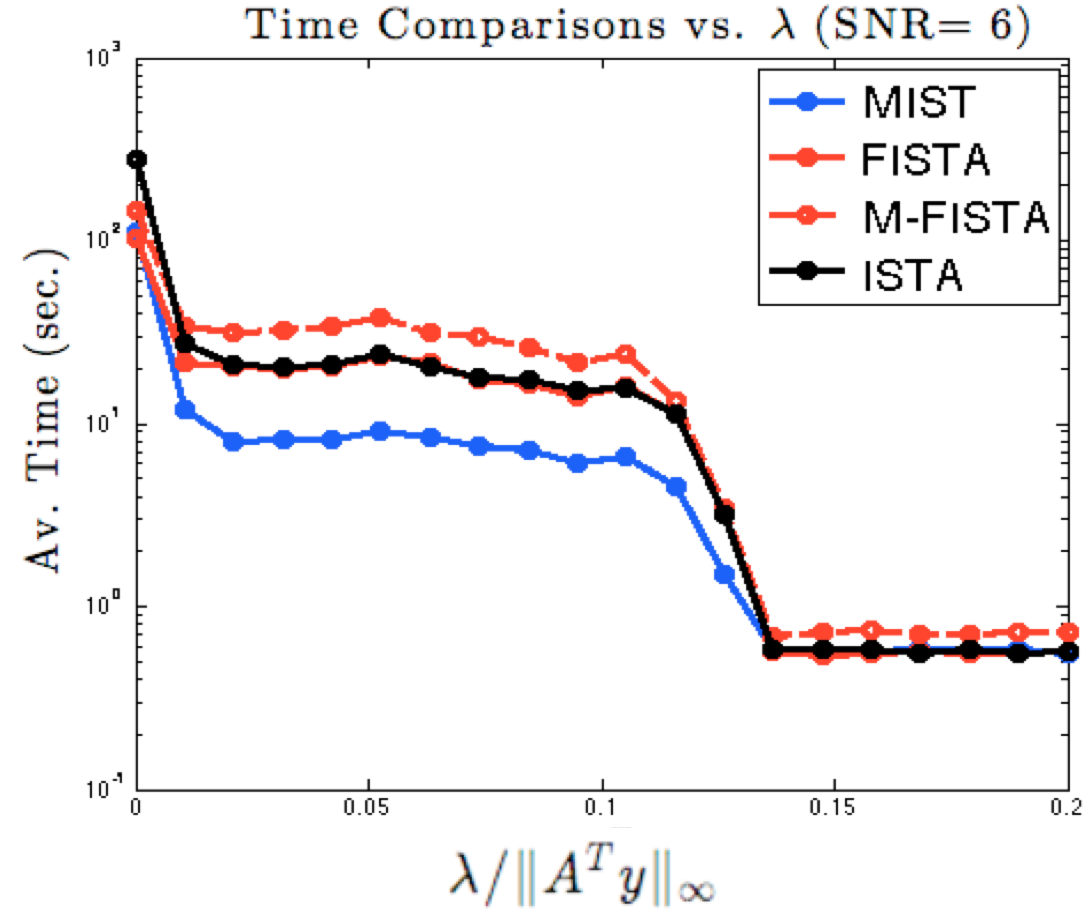}
\end{minipage}%
\caption{Similar comparisons as in Fig. \ref{LamVsTimeSNR12p0 fig} except that SNR=6. As it can be seen, except in the scenario when $\lam=10^{-4}\|\A^T\yv\|_\infty$ the MIST algorithm outperforms the others.}
\label{LamVsTimeSNR6p0 fig}

\vspace{2mm}

\centering
\begin{minipage}{0.7\linewidth}
  \centering
  \includegraphics[width=1.0\linewidth]{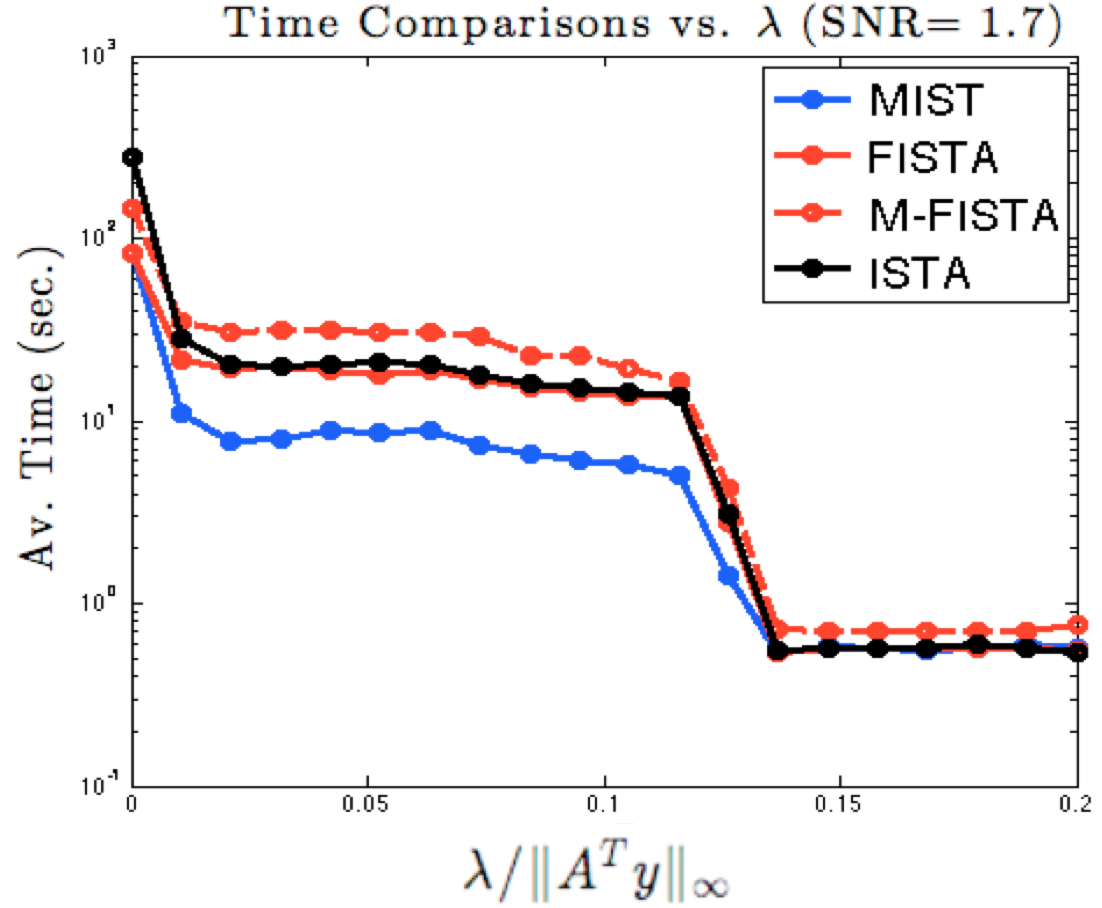}
\end{minipage}%
\caption{Similar comparisons as in Fig. \ref{LamVsTimeSNR12p0 fig} except that SNR=1.7. As it can be seen, except in the scenario when $\lam=10^{-4}\|\A^T\yv\|_\infty$ the MIST algorithm outperforms the others.}
\label{LamVsTimeSNR1p7 fig}

\end{figure}

Lastly, Figures \ref{LamVsTimeSNR12p0 fig}, \ref{LamVsTimeSNR6p0 fig} and \ref{LamVsTimeSNR1p7 fig} show the average speed of each algorithm as a function of $\lam$. 

\section{Conclusion} \label{conclusion section}

We have developed a momentum accelerated MM algorithm, MIST, for minimizing the $l_0$ penalized least squares criterion for linear regression problems. We have proved that MIST converges to a local minimizer without imposing any assumptions on the regression matrix $\A$. Simulations on large data sets were carried out for different SNR values, and have shown that the MIST algorithm outperforms other popular MM algorithms in terms of time and iteration number. 

\section{Appendix} \label{appendix section}

\noindent\textbf{Proof of Theorem \ref{L0 FISTA thm}:} Let $\wv=\xv+\beta\delv$. The quantities $\f(\wv)$, $\gf(\wv)^T(\zv-\wv)$ and $\|\zv-\wv\|_2^2$ are quadratic functions, and by simple linear algebra they can easily be expanded in terms of $\zv$, $\xv$ and $\delv$. Namely,
\begin{align}
f(\wv)&=\frac{1}{2}\|\A\wv-\yv\|^2 \nonumber\\
&=\frac{1}{2}\|\A\xv-\yv+\beta\A\delv\|^2 \nonumber\\
&=f(\xv)+\beta(\A\xv-\yv)^T\A\delv+\frac{1}{2}\beta^2\|\A\delv\|^2 \nonumber\\
&=f(\xv)+\beta\gf(\xv)^T\delv+\frac{1}{2}\beta^2\|\A\delv\|^2. \label{expansion of f}
\end{align}

Also, 

\begin{align}
&\gf(\wv)^T(\zv-\wv) \nonumber\\
&=(\gf(\xv)+\beta\A^T\A\delv)^T(\zv-\wv) \nonumber\\
&=\gf(\xv)^T(\zv-\wv)+\beta\delv^T\A^T\A(\zv-\wv) \nonumber\\
&=\gf(\xv)^T(\zv-\xv-\beta\delv)+\beta\delv^T\A^T\A(\zv-\xv-\beta\delv) \nonumber\\
&=\gf(\xv)^T(\zv-\xv)-\beta\gf(\xv)^T\delv + \beta\delv^T\A^T\A(\zv-\xv) \nonumber\\
&\hspace{0.5cm}-\beta^2\|\A\delv\|^2, \label{expansion of linear term}
\end{align}
and finally:
\begin{align}
\|\zv-\wv\|_2^2&=\|\zv-\xv-\beta\delv\|_2^2 \nonumber\\
&=\|\zv-\xv\|_2^2-\beta\delv^T(\zv-\xv)+\beta^2\|\delv\|_2^2 \label{expansion of quadratic term}
\end{align}
Using the above expansions and the definition of $\Qm(\cdot,\cdot)$, we have that:
\begin{align}
\Qm(\zv,\wv)=\Qm(\zv,\xv)+\Phim(\zv,\delv,\beta), \label{expansion of Q}
\end{align}
where:
\begin{align} 
\Phim(\zv,\delv,\beta)=\frac{1}{2}\beta^2\delv^T\Bm\delv-\beta\delv^T\Bm(\zv-\xv). \nonumber
\end{align}
Observing that $\delv^T\Bm\delv>0$, let:
\begin{align}
\beta=2\eta\left(\frac{\delv^T\Bm(\zv-\xv)}{\delv^T\Bm\delv}\right),\ \eta\in[0,1]. \label{beta expression}
\end{align} 
Then, one has:
\small
\begin{align}
\hspace{-2mm}\Qm(\pmu{\mu}(\wv),\wv)&=\min_{\zv}\Qm(\zv,\wv)\leq\Qm(\zv,\wv) \nonumber\\
&\overset{(\ref{expansion of Q})}{=}\Qm(\zv,\xv)+\Phim(\zv,\delv,\beta) \nonumber\\
&\overset{(\ref{beta expression})}{=}\Qm(\zv,\xv)-2\eta(1-\eta)\frac{[\delv^T\Bm(\zv-\xv)]^2}{\delv^T\Bm\delv} \label{Q inequality} \\
&\leq\Qm(\zv,\xv), \label{Qm inequality}
\end{align}
\normalsize
which holds for any $\zv$. So, letting $\zv=\pmu{\mu}(\xv)$ implies:
\begin{align}
\F(\pmu{\mu}(\wv))\overset{(\ref{Fxkp<=Fxk})}{\leq}\Qm(\pmu{\mu}(\wv),\wv)\overset{(\ref{Qm inequality})}{\leq}\Qm(\pmu{\mu}(\xv),\xv)
\overset{(\ref{Fxkp<=Fxk})}{\leq}\F(\xv), \nonumber
\end{align}
which completes the proof. \qed 

\vspace{2mm}

\noindent\textbf{Proof of Theorem \ref{hard-thresholding thm}:} Looking at (\ref{pmu operator}) it is obvious that:
\begin{align}
\pmu{\mu}(\vv)[i]=\arg\min_{\zv[i]}\frac{1}{2}(\zv[i]-\g(\vv)[i])^2+(\lamu)\Ind(\zv[i]\neq 0). \nonumber
\end{align}
If $|\g(\vv)[i]|\neq\sqrt{2\lamu}$, by \cite[Theorem 1]{MH14} $\pmu{\mu}(\vv)[i]$ is unique and given by $\tmu{\lamu}(\g(\vv))[i]$. If $|\g(\vv)[i]|=\sqrt{2\lamu}$, again by \cite[Theorem 1]{MH14} we now have:
\begin{align}
\pmu{\mu}(\vv)[i]=0 \textup{ and }\pmu{\mu}(\vv)[i]=\sgn(\g(\vv)[i])\sqrt{2\lamu}. \nonumber
\end{align}
Hence, $\tmu{\lamu}(\g(\vv))[i]\in\pmu{\mu}(\vv)[i]$, completing the proof. \qed

\vspace{2mm}

\noindent\textbf{Proof of Lemma \ref{difference in iterates to zero lem}:} From Theorem \ref{L0 FISTA thm}, $0\leq\F(\xkp)\leq\F(\xk)$, so the sequence $\{\F(\xk)\}_{k\geq 0}$ is bounded, which means it has a finite limit, say, $\Fd$. As a result:
\begin{align}
\F(\xk)-\F(\xkp)\to\Fd-\Fd=0. \label{Fk-Fkp limit}
\end{align}
Next, recall that $\wk=\xk+\alpk\delk$ and $\g(\cdot)=(\cdot)-\frac{1}{\mu}\gf(\cdot)$. So, using (\ref{Q inequality}) in the proof of Theorem \ref{L0 FISTA thm} (where $\zv=\pmu{\mu}(\xv)$) with $\wv$, $\pmu{\mu}(\wv)$, $\pmu{\mu}(\cdot)$, $\xv$, and $\delv$ respectively replaced by $\wk$, $\xkp$, $\tmu{\lamu}(\g(\cdot))$, $\xk$ and $\delk$, we have:
\begin{align}
\Qm(\xkp,\wk)&\leq\Qm(\tmu{\lamu}(\g(\xk)),\xk) \nonumber\\
&-2\etak(1-\etak)\frac{[\delk^T\Bm(\tmu{\lamu}(\g(\xk))-\xk)]^2}{\delk\Bm\delk} \nonumber\\
&\leq\F(\xk)-\sigk\alpk^2\delk^T\Bm\delk, \label{Q inequality 2}
\end{align}
where $\sigk=(1-\etak)/2\etak>0$. The first term in (\ref{Q inequality 2}) follows from the fact that:
\begin{align}
\Qm(\tmu{\lamu}(\g(\xk)),\xk)=\Qm(\pmu{\mu}(\xk),\xk)\leq\F(\xk). \nonumber
\end{align} 
Now, noting that:
\begin{align}
\Qm(\xkp,\wk)&=\F(\xkp) \nonumber\\
&+\frac{1}{2}(\xkp-\wk)^T\Bm(\xkp-\wk), \label{Q equality}
\end{align}
which easily follows from basic linear algebra, (\ref{Q inequality 2}) and (\ref{Q equality}) together imply that: 
\begin{align}
\hspace{-2mm}\F(\xk)-\F(\xkp)&\geq \sigk\alpk^2\delk^T\Bm\delk \nonumber\\
&+\frac{1}{2}(\xkp-\wk)^T\Bm(\xkp-\wk) \nonumber\\
&\geq \rho\sigk\alpk^2\|\delk\|_2^2 + \frac{\rho}{2}\|\xkp-\wk\|_2^2, \label{Fk-Fkp inequality}
\end{align}
where $\rho>0$ is the smallest eigenvalue of $\Bm\succ 0$. So, both terms on the right hand side in (\ref{Fk-Fkp inequality}) are $\geq 0$ for all $k$. As a result, due to (\ref{Fk-Fkp limit}) we can use the pinching/squeeze argument on (\ref{Fk-Fkp inequality}) to establish that $\xkp-\wk=\delkp-\alpk\delk\to 0$ and $\alpk\delk\to 0$ as $k\to\infty$. Consequently, $\delk\to 0$ as $k\to\infty$, which completes the proof. \qed

\vspace{2mm}

\noindent\textbf{Proof of Lemma \ref{closed map thm}:} Firstly, from Lemma \ref{difference in iterates to zero lem} we have that $\xkn-\xknm\to 0$ as $n\to\infty$. In the last paragraph in the proof of that lemma (above) we also have that $\alpk\delk\to 0$, and so, $\alpkn\delkn\to 0$ as $n\to\infty$. As a result, by the definition of $\wkn$, $\wkn\to\wvd=\xd$. Then, by the continuity of $\gf(\cdot)$ we have that:
\begin{align}
\wkn-\frac{1}{\mu}\gf(\wkn) \to \xd-\frac{1}{\mu}\gf(\xd). \nonumber
\end{align}
Now, we need to show that, if $\ukn\to\ud$ as $n\to\infty$, then:
\begin{align}
\tmu{\lamu}(\ukn)\to\tmu{\lamu}(\ud). \label{H is closed 3}
\end{align}
Consider an arbitrary component of $\ukn$, say, $\ukn[i]$, in which case we must have $\ukn[i]\to\ud[i]$. Without loss of generality, assume $\ud[i]>0$. Then, by the definition of $\tmu{\lamu}(\cdot)$, there are two scenarios to consider:
\begin{align}
\textup{(a)} \hspace{2mm} \ud[i]\neq\sqrt{2\lamu}  \hspace{10mm} \textup{(b)} \hspace{2mm} \ud[i]=\sqrt{2\lamu}. \nonumber 
\end{align}

Regarding (a): For a large enough $n=N$, we must either have $\ukn[i]<\sqrt{2\lamu}$ or $\ukn[i]>\sqrt{2\lamu}$ for all $n>N$, which implies:
\begin{align}
\tmu{\lamu}(\ukn)[i]=
\begin{cases}
0 & \textup{if } \ukn[i]<\sqrt{2\lamu}  \\[2mm]
\ukn[i] & \textup{if } \ukn[i]>\sqrt{2\lamu},
\end{cases} \label{case (a)}
\end{align}
for all $n>N$. In (\ref{case (a)}), $\tmu{\lamu}(\cdot)$ is a continuous function of $\ukn[i]$ in both cases, which immediately implies (\ref{H is closed 3}).

Regarding (b): In general, $\ud[i]$ could be reached in an oscillating fashion, i.e., for some $n$ we can have $\ukn[i]<\sqrt{2\lamu}$ and for others $\ukn[i]>\sqrt{2\lamu}$. If this was the case for all $n\to\infty$ then $\tmu{\lamu}(\ukn)[i]$ would approach a limit set of two points $\{0,\sqrt{2\lamu}\}$. However, having $\xkn\to\xd$ and $\xkp-\xk\to 0$ implies $\xknp\to\xd$. So, using the fact that:
\begin{align}
\xknp[i]=\tmu{\lamu}(\ukn)[i] \label{xkp=H}
\end{align}
$\tmu{\lamu}(\ukn)[i]$ must approach either $0$ or $\sqrt{2\lamu}$. In other words, there has to exist a large enough $n=N$ such that $\ud[i]$ is approached either only from the left or the right for all $n>N$, i.e., 
\begin{itemize}
\item[(b$_1$)] if $\ukn[i]<\ud[i]=\sqrt{2\lamu}$ for all $n>N$, from (\ref{case (a)}) we have $\tmu{\lamu}(\ukn)[i]\to 0$. So, noting that:
\begin{align}
\tmu{\lamu}(\ukn)[i]-\xkn\overset{(\ref{xkp=H})}{=}\xknp-\xkn\to 0 \label{H-x}
\end{align}
implies $\xkn[i]\to 0$. As a result, $\xd[i]=0$, and using the definition of $\tmu{\lamu}(\cdot)$, we have:
\begin{align}
\tmu{\lamu}(\ud)[i]&=\sqrt{2\lamu}\ \Ind(\wvd[i]\neq 0) \nonumber\\
&=\sqrt{2\lamu}\ \Ind(\xd[i]\neq 0)=\sqrt{2\lamu}\cdot 0 \nonumber\\
&=0. \nonumber
\end{align}
Hence, (\ref{H is closed 3}) is satisfied.

\vspace{1mm}

\item[(b$_2$)] if $\ukn[i]>\ud[i]=\sqrt{2\lamu}$ for all $n>N$, from (\ref{case (a)}) we have $\tmu{\lamu}(\ukn)[i]\to\sqrt{2\lamu}$. So, (\ref{H-x}) implies $\xkn\to\sqrt{2\lamu}$, and using the definition of $\tmu{\lamu}(\cdot)$, we have:
\begin{align}
\tmu{\lamu}(\ud)[i]&=\sqrt{2\lamu}\ \Ind(\wvd[i]\neq 0) \nonumber\\
&=\sqrt{2\lamu}\ \Ind(\xd[i]\neq 0)=\sqrt{2\lamu}\cdot 1 \nonumber\\
&=\sqrt{2\lamu}. \nonumber
\end{align}
Hence, (\ref{H is closed 3}) is again satisfied.
\end{itemize}
Since $i$ is arbitrary, the proof is complete. \qed

\vspace{2mm}

\noindent\textbf{Proof of Lemma \ref{fixed points lem}:} The fixed points are obviously obtained by setting $\xkp=\xk=\xkm=\xd$. So, any fixed point $\xd$ satisfies the equation:
\begin{align}
\xd=\tmu{\lamu}\left(\xd-\frac{1}{\mu}\gf(\xd)\right). \label{fixed point equation}
\end{align}
The result is established by using the definition of $\tmu{\lamu}(\g(\cdot))$ in Theorem \ref{hard-thresholding thm}. Namely, if $i\in\Zo$ we easily obtain:
\begin{align}
|(1/\mu)\gf(\xd)[i]|\leq\sqrt{2\lamu}, \nonumber
\end{align}
which reduces to C$_1$. If $i\in\Zc$, we easily obtain:
\begin{align}
\xd[i]=\xd[i]-(1/\mu)\gf(\xd)[i], \label{equality C2}
\end{align}
which reduces to C$_2$. However, since:
\begin{align}
\left|\xd[i]-(1/\mu)\gf(\xd)[i]\right|\geq\sqrt{2\lamu}, \label{inequality C3}
\end{align} 
(\ref{equality C2}) and (\ref{inequality C3}) together imply $|\xd[i]|\geq\sqrt{2\lamu}$, giving C$_3$. \qed 

\vspace{2mm}

\noindent\textbf{Proof of Lemma \ref{fixed points are local min lem}:} Letting $\Zo=\{i:\xd[i]=0\}$ and $\Zc=\{i:\xd[i]\neq 0\}$, it can easily be shown that $\F(\xd+\dv)=\F(\xd)+\phi(\dv)$, where:
\begin{align}
\phi(\dv)&=\frac{1}{2}\|\A\dv\|_2^2+\dv^T\gf(\xd)+\lam\|\xd+\dv\|_0-\lam\|\xd\|_0 \nonumber\\
&\geq\sum_{i\in\Zo} \underbrace{\dv[i]\gf(\xd)[i]+\lam\Ind(\dv[i]\neq 0)}_{=\phiZo(\dv[i])} \nonumber\\
&+\sum_{i\in\Zc} \underbrace{\dv[i]\gf(\xd)[i] + \lam\Ind(\xd[i]+\dv[i]\neq 0)-\lam}_{=\phiZc(\dv[i])}. \nonumber 
\end{align} 
Now, $\phiZo(0)=0$, so suppose $|\dv[i]|\in\left(0,\lam/\sqrt{2\lam\mu}\right)$, $i\in\Zo$. Then:
\begin{align}
\phiZo(\dv[i])\geq-|\dv[i]||\gf(\xd)[i]|+\lam\geq-|\dv[i]|\sqrt{2\lam\mu}+\lam>0. \nonumber
\end{align}
The second inequality in the above is due to C$_1$ in Lemma \ref{fixed points lem}. 

\vspace{1mm}

Lastly, note that $\phiZc(0)=0$, and suppose $i\in\Zc$. From C$_2$ in Lemma \ref{fixed points lem} we have $\gf(\xd)[i]=0$. Thus, supposing $|\dv[i]|\in(0,\sqrt{2\lamu})$, from C$_2$ we have $|\dv[i]|<|\xd[i]|$ for all $i\in\Zc$. Thus:
\begin{align}
\Ind(\xd[i]+\dv[i]\neq 0)=\Ind(\xd[i]\neq 0)=1, \nonumber
\end{align}
and so, $\phiZc(\dv[i])=0$. Since $\frac{\lam}{\sqrt{2\lam\mu}}<\sqrt{\frac{2\lam}{\mu}}$, the proof is complete after letting $\eps=\frac{\lam}{\sqrt{2\lam\mu}}$. \qed 
 
\vspace{2mm}

\noindent\textbf{Proof of Lemma \ref{limit points are fixed points lem}:} Since it is assumed that $\{\xk\}_{k\geq 0}$ is bounded, the sequence $\{(\xk,\xkp)\}_{k\geq 0}$ is also bounded, and thus, has at least one limit point. Denoting one of these by $(\xd,\xdd)$, there exists a subsequence $\{(\xkn,\xknp)\}_{n\geq 0}$ such that $(\xkn,\xknp)\to(\xd,\xdd)$ as $n\to\infty$. However, by Lemma \ref{difference in iterates to zero lem} we must have $\xkn-\xknp\to 0$, which implies $\xd=\xdd$. Consequently:
\begin{align}
\xknp=\tmu{\lamu}\left(\wkn-\frac{1}{\mu}\gf(\wkn)\right)\to\xd, \label{xknp limit}
\end{align}
recalling that $\wkn=\xkn+\alpkn\delkn$ and $\delkn=\xkn-\xknm$. Furthermore, the convergence:
\small
\begin{align}
\tmu{\lamu}\left(\wkn-\frac{1}{\mu}\gf(\wkn)\right)\to\tmu{\lamu}\left(\xd-\frac{1}{\mu}\gf(\xd)\right), \label{H is closed 2}
\end{align}
\normalsize
follows from Lemma \ref{closed map thm}. Equating the limits in (\ref{xknp limit}) and (\ref{H is closed 2}) assures that $\xd$ satisfies the fixed point equation (\ref{fixed point equation}), making it a fixed point of the algorithm. This completes the proof.  \qed

\vspace{2mm}

\noindent\textbf{Proof of Theorem \ref{convergence thm}:} By Lemma \ref{difference in iterates to zero lem} and Ostrowski's result \cite[Theorem 26.1]{Ostrowski73}, the bounded $\{\xk\}_{k\geq 0}$ converges to a closed and connected set, i.e., the set of limit points form a closed and connected set. But, by Lemma \ref{limit points are fixed points lem} these limit points are fixed points, which by Lemma \ref{fixed points are local min lem} are strict local minimizers. So, since the local minimizers form a discrete set the connected set of limit points can only contain one point, and so, the entire $\{\xk\}_{k\geq 0}$ must converge to a single local minimizer.  \qed

\small
\bibliographystyle{IEEEtran}
\bibliography{Ref}

\begin{thebibliography}{10}
\providecommand{\url}[1]{#1}
\csname url@samestyle\endcsname
\providecommand{\newblock}{\relax}
\providecommand{\bibinfo}[2]{#2}
\providecommand{\BIBentrySTDinterwordspacing}{\spaceskip=0pt\relax}
\providecommand{\BIBentryALTinterwordstretchfactor}{4}
\providecommand{\BIBentryALTinterwordspacing}{\spaceskip=\fontdimen2\font plus
\BIBentryALTinterwordstretchfactor\fontdimen3\font minus
  \fontdimen4\font\relax}
\providecommand{\BIBforeignlanguage}[2]{{%
\expandafter\ifx\csname l@#1\endcsname\relax
\typeout{** WARNING: IEEEtran.bst: No hyphenation pattern has been}%
\typeout{** loaded for the language `#1'. Using the pattern for}%
\typeout{** the default language instead.}%
\else
\language=\csname l@#1\endcsname
\fi
#2}}
\providecommand{\BIBdecl}{\relax}
\BIBdecl

\bibitem{BYD07}
T.~{Blumensath}, M.~{Yaghoobi}, and M.~E. {Davies}, ``Iterative hard
  thresholding and $l_0$ regularisation,'' \emph{IEEE ICASSP}, vol.~3, pp.
  877--880, 2007.

\bibitem{BD208}
T.~{Blumensath} and M.~{Davies}, ``Iterative thresholding for sparse
  approximations,'' \emph{J. Fourier Anal. Appl.}, vol.~14, no.~5, pp.
  629--654, 2008.

\bibitem{MFH11}
R.~{Mazumder}, J.~{Friedman}, and T.~{Hastie}, ``Sparse\uppercase{N}et:
  Coordinate descent with non-convex penalties,'' \emph{J. Am. Stat. Assoc.},
  vol. 106, no. 495, pp. 1--38, 2011.

\bibitem{BLR09}
K.~Bredies, D.~A. Lorenz, and S.~Reiterer, ``Minimization of non-smooth,
  non-convex functionals by iterative thresholding,'' \emph{Tech. Rept.}, 2009,
  \url{http://www.uni-graz.at/~bredies/publications.html}.

\bibitem{Tseng01}
P.~{Tseng}, ``Convergence of block coordinate descent method for
  nondifferentiable minimization,'' \emph{J. Optimiz. Theory App.}, vol. 109,
  no.~3, pp. 474--494, 2001.

\bibitem{Nikolova13}
M.~{Nikolova}, ``Description of the minimisers of least squares regularized
  with $l_0$ norm. uniqueness of the global minimizer,'' \emph{SIAM J. Imaging.
  Sci.}, vol.~6, no.~2, pp. 904--937, 2013.

\bibitem{MS13}
G.~{Marjanovic} and V.~{Solo}, ``On exact $l_q$ denoising,'' \emph{IEEE
  ICASSP}, pp. 6068--6072, 2013.

\bibitem{MS14}
------, ``$l_q$ sparsity penalized linear regression with cyclic descent,''
  \emph{IEEE T. Signal Proces.}, vol.~62, no.~6, pp. 1464--1475, 2014.

\bibitem{BT209}
A.~{Beck} and M.~{Teboulle}, ``A fast iterative shrinkage-thresholding
  algorithm for linear inverse problems,'' \emph{SIAM J. Imaging Sciences},
  vol.~2, no.~1, pp. 183--202, 2009.

\bibitem{WNF09}
S.~J. {Wright}, R.~D. {Nowak}, and M.~{Figueiredo}, ``Sparse reconstruction by
  separable approximation,'' \emph{IEEE T. Signal Proces.}, vol.~57, no.~7, pp.
  2479--2493, 2009.

\bibitem{BT309}
A.~{Beck} and M.~{Teboulle}, ``Fast gradient-based algorithms for constrained
  total variation image denoising and deblurring,'' \emph{IEEE T. Image
  Process.}, vol.~18, no.~11, pp. 2419--2134, 2009.

\bibitem{FBDN07}
M.~Figueiredo, J.~M. Bioucas-Dias, and R.~D. Nowak, ``Majorization-minimization
  algorithms for wavelet-based image restoration,'' \emph{IEEE T. Image
  Process.}, vol.~16, no.~12, pp. 2980--2991, 2007.

\bibitem{Nesterov83}
Y.~{Nesterov}, ``A method of solving a convex programming problem with
  convergence rate $\mc{O}(1/k^2)$,'' \emph{Soviet Math. Doklady}, vol.~27, pp.
  372--376, 1983.

\bibitem{Schwarz78}
U.~J. {Schwarz}, ``Mathematical-statistical description of the iterative beam
  removing technique (method clean),'' \emph{Astron. Astrophys.}, vol.~65, pp.
  345–--356, 1978.

\bibitem{Akaike73}
H.~{Akaike}, ``Information theory and an extension of the maximum likelihood
  principle,'' \emph{2nd International Symposium on Information Theory}, pp.
  267--281, 1973.

\bibitem{Stone74}
M.~{Stone}, ``Cross-validatory choice and assessment of statistical predictions
  (with \uppercase{D}iscussion),'' \emph{J. R. Statist. Soc. B}, vol.~39, pp.
  111--147, 1974.

\bibitem{CW79}
P.~{Craven} and G.~{Wahba}, ``Smoothing noisy data with spline functions:
  \uppercase{E}stimating the correct degree of smoothing by the method of
  generalized cross-validation,'' \emph{Numer. Math.}, vol.~31, pp. 377--403,
  1979.

\bibitem{CC08}
J.~{Chen} and Z.~{Chen}, ``Extended \uppercase{B}ayesian information criteria
  for model selection with large model spaces,'' \emph{Biometrika}, vol.~95,
  no.~3, pp. 759--771, 2008.

\bibitem{BS02}
K.~W. {Broman} and T.~P. {Speed}, ``A model selection approach for the
  identification of quantitative trait loci in experimental crosses,'' \emph{J.
  R. Statist. Soc. B}, vol.~64, pp. 641--656, 2002.

\bibitem{S04}
D.~{Siegmund}, ``Model selection in irregular problems: Application to mapping
  quantitative trait loci,'' \emph{Biometrika}, vol.~91, pp. 785--800, 2004.

\bibitem{BDG04}
M.~{Bogdan}, R.~{Doerge}, and J.~K. {Ghosh}, ``Modifying the \textup{S}chwarz
  \textup{B}ayesian information criterion to locate multiple interacting
  quantitative trait loci,'' \emph{Genetics}, vol. 167, pp. 989--999, 2004.

\bibitem{MH14}
G.~{Marjanovic} and A.~O. {Hero}, ``On $l_q$ estimation of sparse inverse
  covariance,'' \emph{IEEE ICASSP}, 2014.

\bibitem{Ostrowski73}
A.~M. {Ostrowski}, \emph{Solutions of Equations in \uppercase{E}uclidean and
  \uppercase{B}anach Spaces}.\hskip 1em plus 0.5em minus 0.4em\relax New York:
  Academic Press, 1973.

\end{thebibliography}

\end{document}